\documentclass[conference,9pt]{IEEEtran}
\IEEEoverridecommandlockouts
\usepackage{cite}
\usepackage{amsmath,amssymb,amsfonts}
\usepackage{algorithmic}
\usepackage{graphicx}
\usepackage{textcomp}
\usepackage{pifont}
\usepackage{tabularx}
\usepackage{booktabs} 
\usepackage{adjustbox}
\usepackage{multirow}
\usepackage{xcolor}
\usepackage{subcaption}
\def\BibTeX{{\rm B\kern-.05em{\sc i\kern-.025em b}\kern-.08em
    T\kern-.1667em\lower.7ex\hbox{E}\kern-.125emX}}
    
\begin{document}

\title{Physics-aware Roughness Optimization for Diffractive Optical Neural Networks}

\author{Shanglin Zhou*$^1$, Yingjie Li*$^2$, Minhan Lou$^2$, Weilu Gao$^2$, Zhijie Shi$^1$, Cunxi Yu$^1$, Caiwen Ding$^1$\\ $^1$University of Connecticut, $^2$University of Utah 
\\
*Both authors contributed equally to this research
}


\maketitle

\begin{abstract}
As a representative next-generation device/circuit
technology beyond CMOS, diffractive optical neural networks (DONNs) have shown promising advantages over conventional deep neural networks due to extreme fast computation speed (light speed) and low energy consumption.
However, there is a \textit{mismatch}, i.e., significant prediction accuracy
loss, between the DONN numerical modelling and physical optical device deployment,
because of the interpixel interaction within the diffractive layers.
In this work, we propose a physics-aware diffractive optical neural network training framework to reduce the performance difference between numerical modeling and practical deployment. Specifically, we propose the roughness modeling regularization in the training process and integrate the physics-aware sparsification method to introduce sparsity to the phase masks to reduce sharp phase changes between adjacent pixels in diffractive layers. We further develop $2\pi$ periodic optimization to reduce the roughness of the phase masks to preserve the performance of DONN. 
Experiment results demonstrate that,
 compared to state-of-the-arts, our physics-aware optimization can provide $35.7\%$, $34.2\%$, $28.1\%$, and $27.3\%$ reduction in roughness with only accuracy loss on MNIST, FMNIST, KMNIST, and EMNIST, 
respectively.

\end{abstract}


\section{Introduction}
\label{intro}

The high computation and memory storage of deep neural networks (DNNs) pose intensive challenges to the conventional Von-Neumann architecture~\cite{islam2022eve, peng2021binary, islam2022enabling}, introducing substantial data movements in memory hierarchy. The yearning for the ultra-efficient DNN accelerators has driven the studies on many different Von-Neumann architectures. We are in urgent need of (i) a next-generation device/circuit technology beyond CMOS and (ii) the corresponding customized algorithm-technology co-design to achieve ultra-low power, real-time DNN processing in various applications.

As a representative, diffractive optical neural networks (DONNs) has drawn much attention. DONNs realizes the all-optical processing based on the physical phenomena, i.e., light diffraction and light signal phase modulation, which happen by nature at the light speed. Weights are encoded as complex-valued transmission coefficients in diffractive layers, and the free-space propagation function~\cite{ersoy2006diffraction} is adopted to multiply with the light wave function to realize computation. Multiple layers of diffractive surfaces features millions of neurons and physically formed the neural network, which mimic the morphology of artificial neural network, while consuming significantly less energy compared to running conventional DNNs on digital platforms.
Therefore, we could achieve high system throughput and computation speed,
with no extra energy cost required for maintaining the function of the computation units (pixels in the phase mask) in DONN systems once the phase masks are fabricated or assembled~\cite{lin2018all, zhou2021large, li2021late, li2022physics, chen2022physics}.

Current DONN systems for all-optical inference is trained on digital platforms with the numerical modelling of the DONN system. However, there is a \textit{mismatch}, i.e., significant prediction accuracy degradation, between numerical modelling and physical deployment.
Potential reasons of such mismatch are:
discrete control levels in optical devices~\cite{zhou2021large}, fabrication errors of devices~\cite{li2022rubikonns, chen2022physics, chen2022complex}, inaccurate emulation kernels~\cite{chen2022physics}, etc. Among them, we identify the interpixel crosstalk \cite{yariv1993interpage} within the phase mask as most critical~\cite{lou2023effects,vallone2018diffusive,o2004diffractive}\,
since it breaks down the emulated computed optical responses from numerical modelling, as the sharp changes between adjacent pixels will introduce a fast-varying incident field. Thus, the impacts of interpixel crosstalk can be quantified using adjacency pixel thickness differences, namely \textit{roughness}. 

To narrow the accuracy gap between digital emulation and hardware deployment, in this paper, we propose a physics-aware roughness optimization process for the DONN system.
Specifically, our proposed physics-aware optimization process includes three steps (Table \ref{tbl:pruning_feature}): \textbf{(1)} integrating roughness regularization into DONN loss function during training. Details are illustrated in Section \ref{roughness}, where a regularization factor is introduced in the training loss for the roughness-aware DONN training; \textbf{(2)} compressing models with block sparsification,
which migrates interpixel interaction by leaving more space between active pixels (Section \ref{blk_prune}); \textbf{(3)} further smoothing masks with $2\pi$ periodic phase modulation. 
Our work outperforms existing approaches~\cite{lin2018all,zhou2021large,li2022physics} for two reasons. First, \cite{lin2018all} did not model the interpixel crosstalk in DONN, therefore is roughness oblivious. Our work, however, integrates the interpixel crosstalk impact into the DONN loss function to be accuracy awareness \cite{lou2023effects}.
Second, previous studies~\cite{zhou2021large,li2022physics} leverage the periodic characteristic of phase modulation for deploying negative phase parameters on existing hardware, i.e., phase modulation is physically equivalent by adding $2\pi$ or its multiple. However, this periodic characteristic has not yet being explored in reducing interpixel crosstalk impacts.


\begin{table}[t]
\caption{Comparison of different methodologies.}
\label{tbl:pruning_feature}
\centering
\begin{adjustbox}{width=0.93\columnwidth, center}
\begin{tabular}{c|ccc}
\toprule
\textbf{Methods} & \multicolumn{1}{c}{\bf Roughness-aware} & \multicolumn{1}{c}{\bf Sparsity} & \multicolumn{1}{c}{\begin{tabular}[c]{@{}c@{}}{\bf 2$\pi$~Periodic} \\ {\bf Optimization}\end{tabular}} \\ 
\toprule
\cite{lin2018all,mengu2020scale}                  & \ding{53}  & \ding{53}  &  \ding{53} \\ 
\midrule
\cite{zhou2021large,li2022physics} & \ding{53}  & \ding{53}  & \checkmark \\ 
\midrule
Ours                               & \checkmark & \checkmark & \checkmark \\ 
\toprule
\end{tabular}
\end{adjustbox}
\end{table}


Our contributions are summarized as follows:
\begin{itemize}
\item We propose a physics-aware roughness optimization framework, which integrates roughness modeling and weight sparsification into DONN training process, to effectively narrow the performance mismatch between digitally emulated DONN and physical hardware deployment.
    
\item We propose a roughness modeling method, 
which smooths the phase masks in DONNs training and also provides a method to quantify the performance gap.

\item We leverage the physics-aware block sparsification method into the optimization process to introduce sparsity of the DONN system and reduce the
interpixel interactions.

\item We apply $2\pi$ periodic phase modulation to further smooth the phase masks and reduce the performance mismatch.
\end{itemize}

Evaluation results demonstrate that our proposed physic-aware optimization framework provides up to $35.7\%$ 
reduction in roughness.
To the best of our knowledge, this is the first work to address the performance \textit{mismatch} between digital emulation and hardware deployment of DONNs.
 

\begin{figure*}[!ht]
    \centering
    \includegraphics[width=\linewidth]{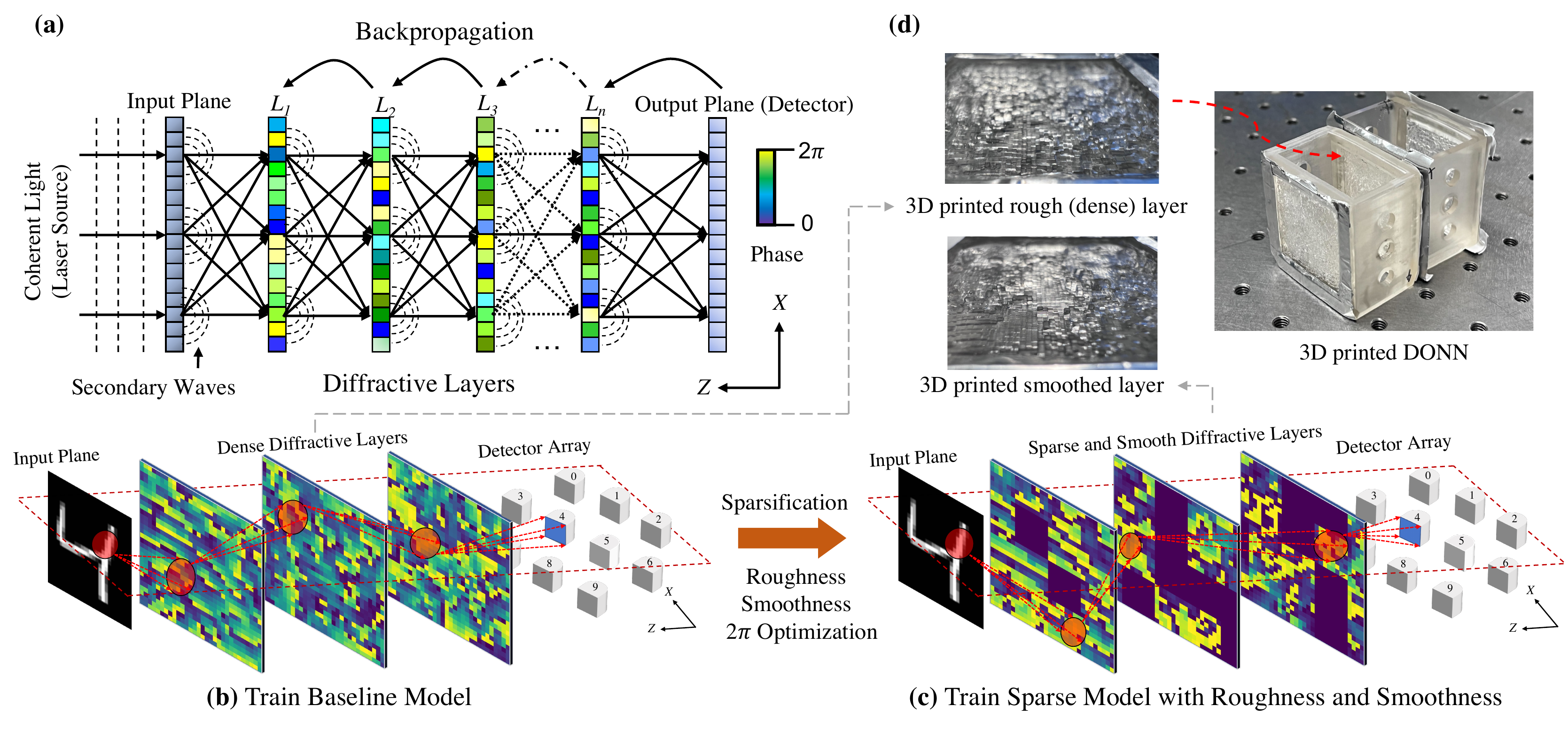}
    \vspace{-0.6cm}
    \caption{Overview of the proposed physics-aware roughness optimization process. (a) Illustration of the DONN system including the input plane, multiple diffractive layers, and a detector (output) plane. (b) Illustration of the training process of baseline model. (c) Illustration of the roughness-aware sparsification process. The black blocks in the three diffractive layers are sparsified with all values in them are zero. (d) Images of 3D printed dense and smoothed layer, and DONN. 
    }
    \label{fig:donn_system}
\vspace{-0.5cm}
\end{figure*}


\section{Background and Related Works}

\subsection{Diffractive Optical Neural Networks (DONNs)}



There are two aspects contributing to the low-carbon footprint by DONN systems: (1) The propagation of information-encoded light signal happens by nature in physics, i.e., the data movements happen at no additional cost; (2) DONN systems are usually implemented with passive devices, such as 3D printed phase mask (shown in Fig. \ref{fig:donn_system}(d)), non-violated liquid crystal array, phase-change-materials, which means once the devices are fabricated and assembled, they will provide phase modulation to the light signal at no extra energy cost, i.e., the computation also happens at no cost.

There are two sets of parameters in the DONNs. The first set (trainable) is for phase modulation provided at each diffraction pixel in diffractive layers. The different phases of the input light waves can result in different light intensity distributions at the end of the system.
The second set (non-trainable) is for diffraction approximation. This describes the propagation of the light wave with diffraction at each diffractive layer, which connects the neurons between the layers. 
This set of parameters
is computed by the mathematical approximation for real-world physical phenomena~\cite{li2022physics}. 
The final output of the DONN system, e.g., the predicted label under classification task, can be expressed as the maximum energy value of the last layer's output light intensity that observed by detectors. Hence, similar to the training process of conventional DNNs, optimal weights for phase modulation in diffractive layers in DONN can be obtained by minimizing the commonly used machine learning loss function~\cite{lin2018all}.

\vspace{-0.07cm}
\subsection{DONN Hardware Deployment Challenges}
\label{background:deploy}

The deployment of the digitally trained DONN model on practical optical devices can introduce significant performance degradation~\cite{zhou2021large}, especially with the subwavelength structure~\cite{mansouree2021large, skarda2022low}.
The interpixel interaction within phase masks (diffractive layers) can result in fast-varying incident field, which breaks down the computed optical responses,
resulting the significant miscorrelation gap between numerical modelling and hardware deployment of the DONN system. For example, Zhou et al.\cite{zhou2021large} claims $\geq 30\%$ accuracy degradation while deploying the model to the physical optical system directly without considering the mask roughness.



\subsection{Weight Sparsification}


Weight sparsification has been studied to reduce the model size and accelerate the computation~\cite{ zhang2018systematic, zhou2021end, zhou2022combining}. However, in this paper, we aim to use the sparsification technique to optimize the weight distribution~\cite{zhang2021unified} and force the smoothness of the phase masks in the DONN.


Non-structured magnitude weight sparsification~\cite{han2015learning} has been proven to not only enable the high sparsity of the model but also maintain accuracy. However, the non-zero elements in the sparse weight matrix are randomly distributed after the non-structured sparsification~\cite{han2016eie, wen2016learning}.
This leads to extra effort for the design of the voltage control pattern because of the irregular distribution of weights (phase modulations) on diffractive layers when deploying the DONN system to the physical optical devices.
Structured sparsification, such as bank-balanced sparsification~\cite{yao2019balanced, cao2019efficient} and block-circulant matrix sparsification~\cite{ding2017circnn} has been developed with more effort on the sparsification patterns and provides higher regularity on non-zero elements in the weight matrix. However, these methods still focus on element-wise patterns and do not optimize the distribution and roughness of the weight, which do not provide good smoothness of the weight matrix of the neural networks.

\section{Physics-aware Roughness Optimization}

We migrate the mismatch between the numerical modelling of DONN and practical deployment w.r.t interpixel interaction from two aspects: 
\textbf{(1)} Smooth the phase mask, i.e., reduce the sharp phase changes between neighboring pixels. By reducing the roughness of the mask, the incident field varies smoothly, which maintains its consistency with the optical response computed in training. 
\textbf{(2)} Reduce functioning diffraction pixels. We reduce less important pixels to exact zeros in diffractive layers makes the phase mask more sparse and leaves more space between remaining pixels, thus reducing the interpixel interactions. See Fig.~\ref{fig:donn_system} (b-c) for the overall picture.

\subsection{Differentiable Modeling of DONNs}
\label{onn}

There are three components (Fig. \ref{fig:donn_system}(a)) in a DONN system: (1) laser source for input images encoding, (2) diffractive layers for providing trainable phase modulation, and (3) detectors for capturing the diffraction pattern of the forward propagation. Specifically, the input image is first encoded with the coherent laser light. The information-encoded light signal is diffracted in the free space between diffractive layers, and modulated via phase modulation at each layer. Finally, the diffraction pattern after light propagation w.r.t light intensity distribution is captured at the detector plane for predictions. 

The input (e.g., an image) is encoded on the coherent light signal from the laser source, its wavefunction can be expressed as $f_{0}(x_{0}, y_{0})$. The wavefunction after light diffraction over diffraction distance $z$ to the first diffractive layer
is the summation of the outputs at the input plane, 
i.e., 
\begin{equation}
\label{eq:diffraction_time}
    f_{1}(x, y) = \iint f_{0}(x_{0}, y_{0})h(x-x_{0}, y-y_{0}, z)dx_{0}dy_{0}
\end{equation}
where $(x, y)$ is the coordinate on the receive plane. 
$h$ is the impulse optical response function of free space, which is the mathematical approximation for light diffraction, e.g., Rayleigh-Sommerfeld approximation, Frensel approximation, Frauhofer approximation~\cite{tobin1997introduction}.

 Equation \ref{eq:diffraction_time} can be calculated with spectral algorithm, where we employ Fast Fourier Transform (FFT) for fast and differentiable computation, i.e., 
    $U_{1}(\alpha, \beta) = U_{0}(\alpha, \beta)H(\alpha, \beta, z)$,
where $U$ and $H$ are the Fourier transformation of $f$ and $h$ respectively. 
The
wavefunction
$U_{1}(\alpha, \beta)$ is transformed to time domain with inverse FFT (iFFT) for phase modulation:
    $f_{2}(x, y) = \text{iFFT}(U_{1}(\alpha, \beta)) \times \mathbf{W}_{1}(x, y)$,
where $\mathbf{W}_{1}(x, y)$ is the phase modulation in the first diffractive layer. 
$f_{2}(x, y)$ is the 
wavefunction for the light diffraction for the second diffractive layer. 

We wrap one computation round of light diffraction and phase modulation at one diffractive layer as a computation module named \textbf{DiffMod}, i.e.,
    $\text{DiffMod}(f(x, y), \mathbf{W}) = L(f(x, y), z) \times \mathbf{W}(x, y)$,
where $f(x, y)$ is the input wavefunction.
$\mathbf{W}(x, y)$ is the phase modulation. 
$L(f(x, y), z)$ is the wavefunction after light diffraction over a constant distance $z$ in time domain.  
For a 3-layer DONN system,
the forward function can be expressed as,
\begin{equation}
\label{eq:forward}
\small
\begin{split}
I(f_{0}(x, y), \mathbf{W}) = \text{DiffMod}(\text{DiffMod}(\text{DiffMod}(f_{0}(x, y), \\
\mathbf{W}_{1}(x, y)),\mathbf{W}_{2}(x, y)), \mathbf{W}_{3}(x, y))
\end{split}
\end{equation}
The final diffraction pattern w.r.t the light intensity $I$ in Equation \ref{eq:forward} is captured at the detector plane. 
There are pre-defined detector regions
to mimic the output of conventional neural networks for class prediction. The class with the highest sum of light intensity within its corresponding detector region produced by \texttt{argmax} is picked as the prediction result~\cite{lin2018all,li2020multi}.
Suppose
the ground truth class $t$, the loss function $\boldsymbol\ell$
using \textbf{MSELoss} is
    $\boldsymbol\ell = \parallel \text{Softmax}(I) - t \parallel_{2}$.
Thus, the whole system is designed to be differentiable and compatible with conventional automatic differential engines.


\subsection{Roughness Modeling}
\label{roughness}
\begin{figure}[b]
     \centering
     \begin{subfigure}[b]{0.49\linewidth}
         \centering       \includegraphics[width=0.45\linewidth]{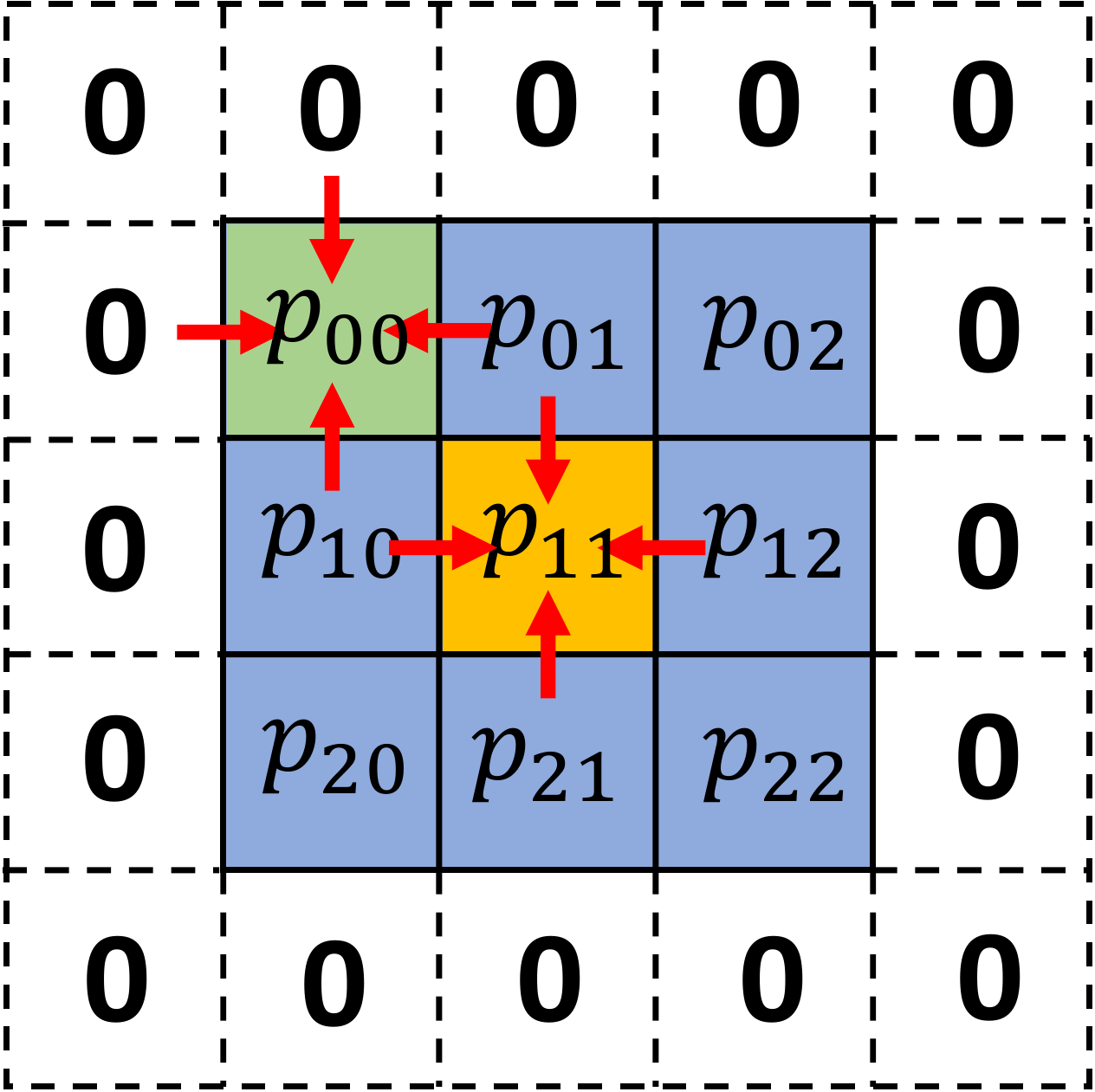}
         \caption{
         4-neighbors.}
         \label{fig:rough_4}
     \end{subfigure}
     \hfill
     \begin{subfigure}[b]{0.49\linewidth}
         \centering
      \includegraphics[width=0.45\linewidth]{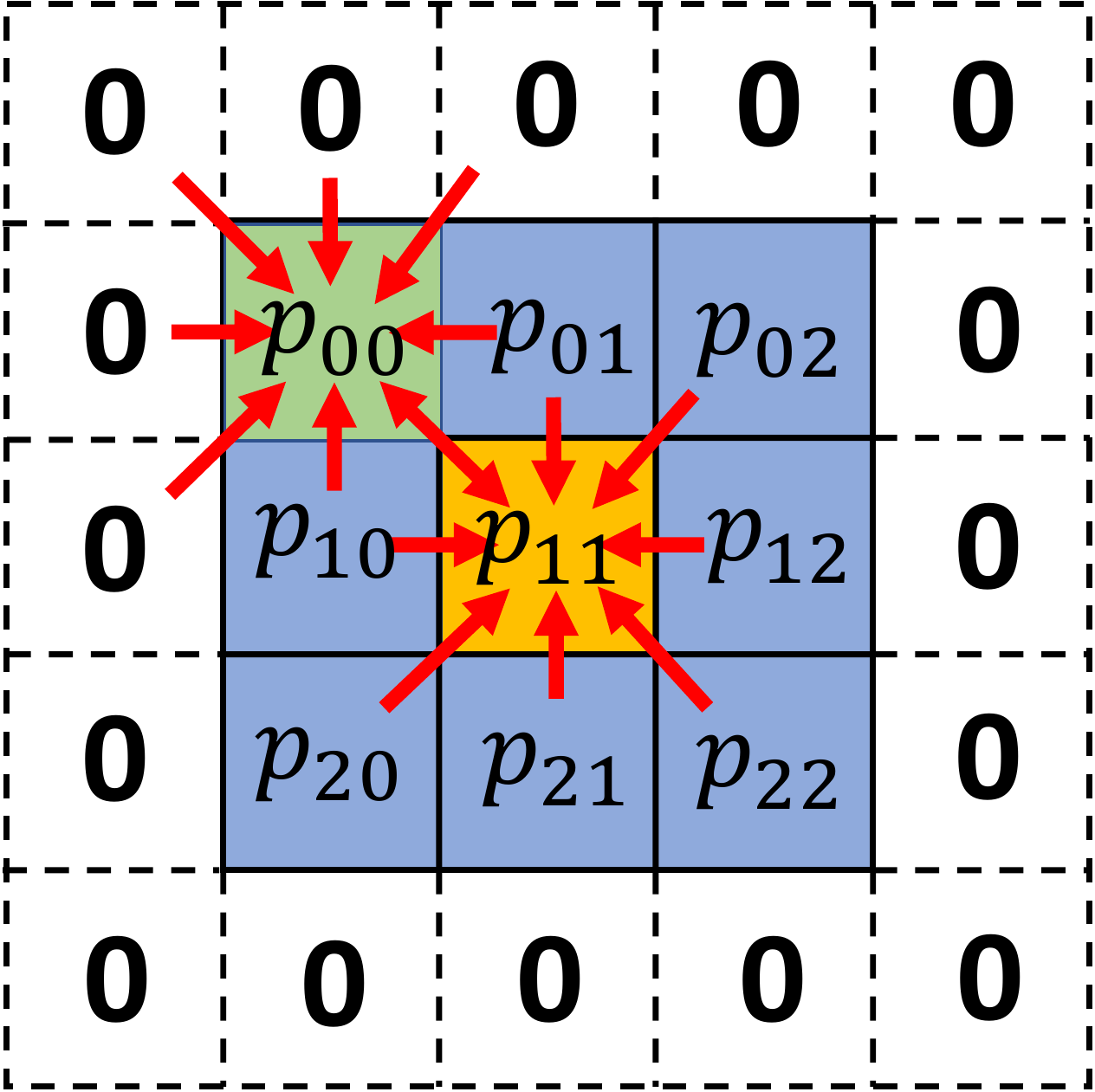}
         \caption{
         8-neighbors.}
         \label{fig:rough_8}
     \end{subfigure}
        \caption{Illustration of roughness modelling for a phase mask with size of $3\times3$ -- (a) with 4 neighbors, (b) with 8 neighbors. }
        \label{fig:rough}
\end{figure}



We develop the the roughness modelling applicable independently for each diffractive layer and differentiable with the training process, targeting the roughness-aware training of the DONN model.
As shown in Fig. \ref{fig:rough}, the roughness of the pixel is computed with the values of its neighboring pixels. 
The one-dimension zeros padding is applied to the original phase mask for roughness computation of pixels at the boundary.
We average the $L_2$-norm differences of value at the pixel and values of its neighbors as the roughness at the pixel. For example, the roughness of the pixel $p_{11}$ in Fig. \ref{fig:rough}, the $k$-neighbor roughness is computed as
\begin{equation}
\small
    R(p_{11}) = \frac{1}{k} \times \Sigma_{ij} \parallel p_{ij} - p_{11} \parallel_{2}, p_{ij} \in \text{neighbor}(p_{11}) 
\end{equation}
where $k$ can be 4 or 8.
For 4-neighbors (Fig. \ref{fig:rough_4}), $\text{neighbor}(p_{11}) = \{p_{01}, p_{10}, p_{12}, p_{21}\}$; and for 8-neighbor roughness (Fig. \ref{fig:rough_8}), $\text{neighbor}(p_{11}) = \{p_{00}, p_{01}, p_{02}, p_{10}, p_{12}, p_{20}, p_{21}, p_{22}\}$.
The roughness of the whole phase mask can be computed with the summation of the roughness of each pixel, i.e., 
\begin{equation}
\label{rough_score}
\small
    R(\mathbf{W}) = \Sigma_{ij} R(p_{ij}), i\in[0, N-1], j\in[0, N-1]
\end{equation}
where the phase mask $\mathbf{W}$ has the size of $N\times N$.
We sum up the roughness of all pixels as the roughness score of the diffractive layer.
We further integrate the roughness matrix
into training,
i.e., 
\begin{equation}
\label{rough_loss}
\small
    L = \parallel \text{Softmax}(I) - t \parallel_{2} + p \times R(\mathbf{W}) 
\end{equation}
where $p$ is the regularization factor and $R(\mathbf{W})$ is the roughness results for each diffractive layer.



\subsection{Physic-aware sparsification: Block sparsification}
\label{blk_prune}

\begin{figure}[t]
\centering
  \includegraphics[width=0.9\linewidth]{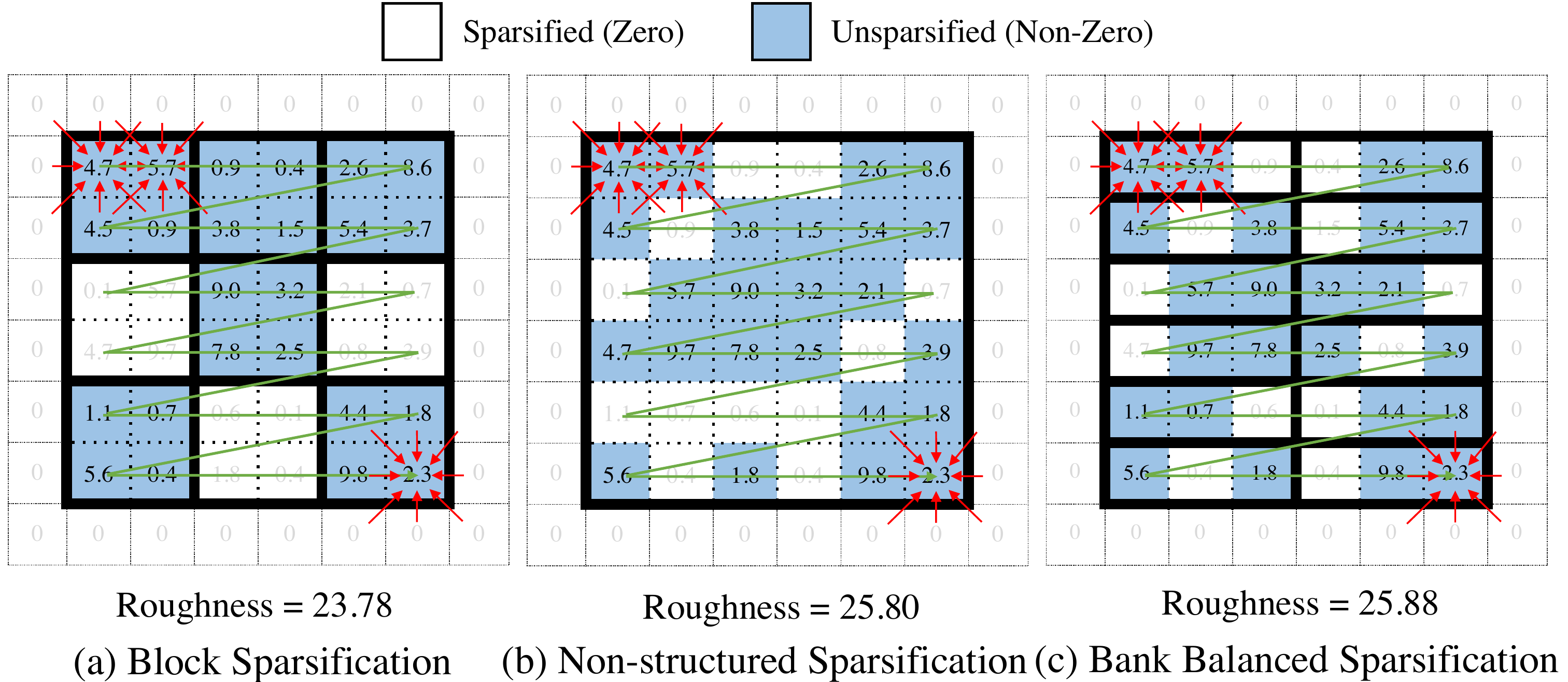}
\vspace{-0.1cm}
\caption{Comparison of block sparsification with two other sparsification methods. All three are with sparsification ratio being 0.33.}
\label{prune}
\vspace{-0.1cm}
\end{figure}

\vspace{-0.1cm}
\subsubsection{Roughness-Aware Sparsification}

To consider the case of different sparsification methods (i.e., bank balanced sparsity, irregular sparsity, and block sparsity), Fig.~\ref{prune} shows the roughness impact using same sparsity ratio (i.e., $0.33$).
Roughness is computed based on Equation~\ref{rough_score} with $8$-neighbors.
For the block sparsification (Fig.~\ref{prune}(a)), we partition the weight matrix into equal-sized blocks. 
Based on the pre-defined threshold or percentile, all the weights within the blocks with the $L_2$-norm smaller than the threshold or the percentile are set to zero. 
For the non-structured sparsification (Fig.~\ref{prune} (b)), weights with absolute value that is smaller than the pre-defined threshold or percentile are sparsified~\cite{han2015learning}. For the bank-balanced sparsification (Fig.~\ref{prune}(c)), rows of weight matrix are split into equal-sized banks, and identical sparsity among banks is kept while sparsification~\cite{cao2019efficient}.
Among all, block sparsification focuses more on the whole blocks of the weight rather than the element-wise pattern, which leads to the lowest roughness.

\subsubsection{SLR-Regularized Optimizations}
We integrate the block weight sparsification method~\cite{lagunas2021block} with the Surrogate Lagrangian Relaxation (SLR)-based model compression optimization technique~\cite{gurevin2020enabling} to reduce functioning diffraction pixels and further achieve smoothness of the phase masks in DONN (Fig.~\ref{fig:donn_system}(c)).
For a DONN with $N$ layers, as $i \in 1,2,...,N$, denote the weights of each diffractive layer as $\mathbf{W}_i$. The objective of our block sparsification is to minimize the DONN loss function (Equation~\ref{rough_loss}) while reducing the number of non-zero blocks of weights in each $\mathbf{W}_i$, i.e., 
$\min_{\mathbf{W}_i} \{\boldsymbol\ell(\mathbf{W}_i) + 
\boldsymbol\ell_{r}(\mathbf{W}_i)\}$
subject to \ding{172} \# non-zero block rows in $\mathbf{W}_i$ is less than $r_i$ and \ding{173} \# non-zero block columns in $\mathbf{W}_i$ is less than $c_i$, where $r_i$ and $c_i$ are the desired non-zero block of rows and columns respectively, and $i = 1, ..., N$.
The unconstrained form can be written as Equation~\ref{relax}.
\vspace{-0.3cm}
\begin{equation}
\small
\begin{aligned}
\label{relax}
&\min_{\mathbf{W}_i}\boldsymbol\ell(\mathbf{W}_i) + 
\boldsymbol\ell_{r}(\mathbf{W}_i)
+ \sum_{i=1}^N g_i\left(\mathbf{W}_i\right) \\
\text{where} &\;
g_{i}\left(\mathbf{W}_{i}\right)=\left\{\begin{array}{ll}
0 & \text { if \ding{172} and \ding{173} are satisfied}  \\
+\infty & \text { otherwise }
\end{array}\right.
\end{aligned}
\vspace{-0.2cm}
\end{equation}
$\boldsymbol\ell(.) + 
\boldsymbol\ell_{r}(.)$ represents the DONN system loss with roughness regularization. $g_i(.)$ is the indicator function that represents the non-differentiable penalty term for each diffractive layer. To solve this,
%
 the duplicate variables $\mathbf{Z}_i$ is introduced~\cite{boyd2011distributed,gurevin2020enabling}.
 The loss function is equivalent as: $\min_{\mathbf{W}_i} \{\boldsymbol\ell(\mathbf{W}_i) + 
\boldsymbol\ell_{r}(\mathbf{W}_i) + \sum_{i=1}^N g_{i}(\mathbf{Z}_i)\}, \: s.t. \: \mathbf{W}_i = \mathbf{Z}_i, i = 1, ..., N$. The resulting Augmented Lagrangian function is 
\vspace{-0.2cm}
\begin{equation}
\label{Aug}
\small
\begin{aligned}
\mathcal{L} \left(\mathbf{W}_i, \mathbf{Z}_i, \mathbf{\Lambda}_i\right) & = 
\boldsymbol\ell(\mathbf{W}_i) + 
\boldsymbol\ell_{r}(\mathbf{W}_i) + \sum_{i=1}^N g_i\left(\mathbf{Z}_i\right) \\
&+\sum_{i=1}^N \operatorname{tr}\left[\mathbf{\Lambda}_i^T\left(\mathbf{W}_i-\mathbf{Z}_i\right)\right]+\sum_{i=1}^N \frac{\rho}{2}\left\|\mathbf{W}_i-\mathbf{Z}_i\right\|_F^2
\end{aligned}
\vspace{-0.2cm}
\end{equation}
Here, $\mathbf{\Lambda}_i$ are Lagrangian multipliers for relaxing the constraints. Their violations are penalized by quadratic penalties with coefficient $\rho$. $\|\cdot\|_F$ is the Frobenius norm. $tr(.)$ is the trace.
We could decompose the above problem into
two subproblems and solve them iteratively until convergence. At iteration $k$, the first subproblem is: given $\mathbf{\Lambda}_i^k$ and $\mathbf{Z}_i$ from the previous iteration, solving the loss function $\min_{\mathbf{W}_i}\mathcal{L}\left(\mathbf{W}_i, \mathbf{Z}_i^{k-1}, \mathbf{\Lambda}_i^k\right)$. At this time, we check the surrogate optimality condition $\mathcal{L}\left(\mathbf{W}_i^k, \mathbf{Z}_i^{k-1}, \mathbf{\Lambda}_i^k\right)<\mathcal{L}\left(\mathbf{W}_i^{k-1}, \mathbf{Z}_i^{k-1}, \mathbf{\Lambda}_i^k\right)$. If 
satisfied, $\mathbf{\Lambda}_i^k$ would be updated as $\mathbf{\Lambda^\prime}_i^{k} = \mathbf{\Lambda}_i^k+s^{\prime k}\left(\mathbf{W}_i^k-\mathbf{Z}_i^{k-1}\right)$. The second subproblem is: fixing $\mathbf{W}_i^k$ from the first subproblem, solving the loss function $\min_{\mathbf{Z}_i}\mathcal{L}\left(\mathbf{W}_i^k, \mathbf{Z}_i, \mathbf{\Lambda^\prime}_i^{k}\right)$. Similar,  we need to check the surrogate optimality condition $\mathcal{L}\left(\mathbf{W}_i^k, \mathbf{Z}_i^k, \mathbf{\Lambda^\prime}_i^{k}\right) < \mathcal{L}\left(\mathbf{W}_i^k, \mathbf{Z}_i^{k-1}, \mathbf{\Lambda^\prime}_i^{k}\right)$ to update $\mathbf{\Lambda}_i^{k+1} = \mathbf{\Lambda^\prime}_i^k+s^k\left(\mathbf{W}_i^k-\mathbf{Z}_i^{k}\right)$. Both $s^{\prime k}$ and $s^{k}$ are stepsize~\cite{gurevin2020enabling}.


\subsection{Smoothness}
\label{smooth}

\subsubsection{Intra-block Smoothness}
\label{block_smooth}
\begin{figure}[t]
\centering
  \includegraphics[width=0.6\linewidth]{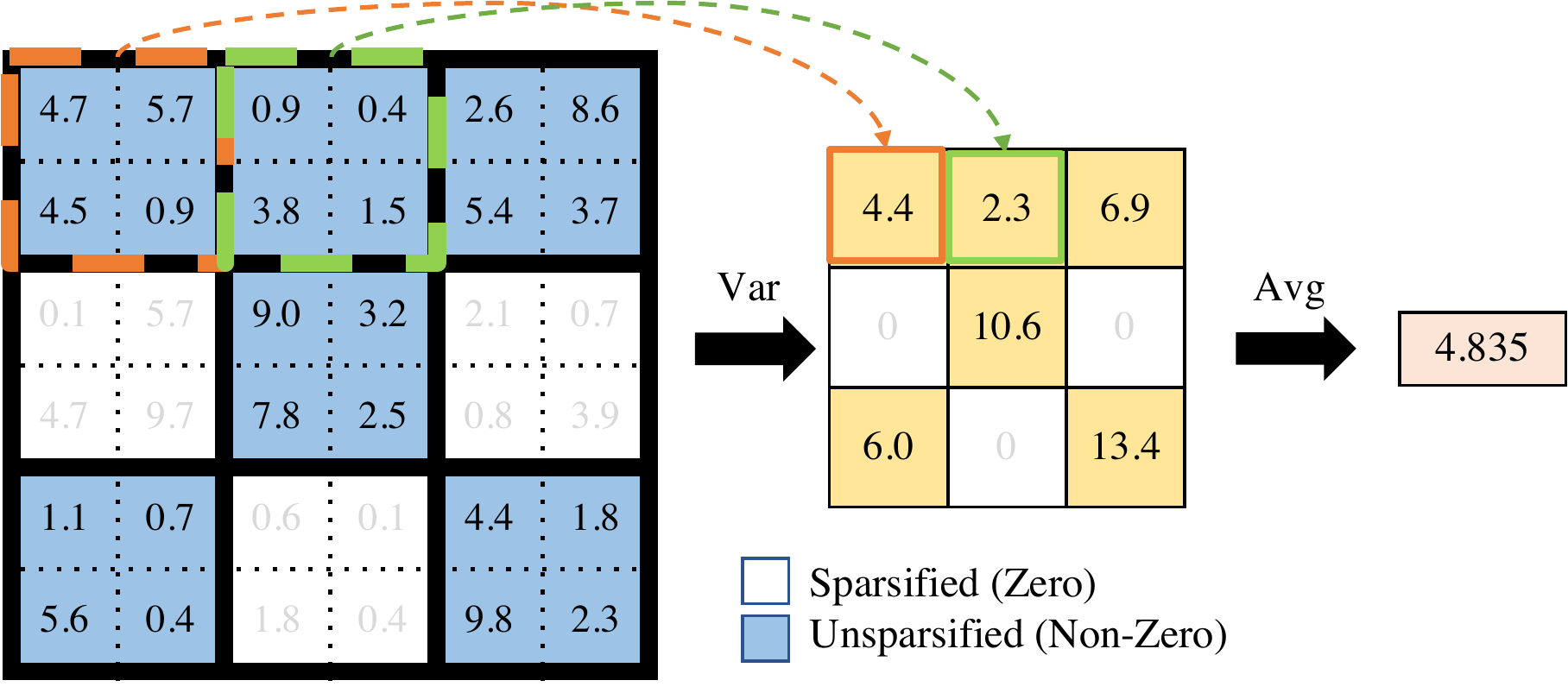}
\caption{Illustration of intra-smoothness for a sparsified phase mask with sparsification ratio being $0.33$, block size being $2$.}
\label{intra_rough}
\end{figure}

Roughness modeling pushes the smoothness of the entire weight matrix in diffractive layers during training, while block sparsification partitions the weight matrix into blocks with all the weights in the sparsified blocks being set to zero and the weights in the unsparsified blocks being distributed irregularly.
To reduce the sharp phase changes in the unsparsified blocks, we perform intra-block smoothness, again with the premise that it is differentiable with the training process. As the example shown in Fig.~\ref{intra_rough}, the weight matrix of a diffractive layer is split into nine $2 \times 2$ blocks. Variance of each block is computed and summed together. The obtained summation is added to the DONN system loss function as the roughness as Equation~\ref{eq:intra_rough}.
\vspace{-0.1cm}
\begin{equation}
\label{eq:intra_rough}
\small
L = \parallel \text{Softmax}(I) - t \parallel_{2} + p \times R(\mathbf{W}) + q \times R_{intra}(\mathbf{W})
\vspace{-0.1cm}
\end{equation}
where $q$ is the regularization factor and $R_{intra}(\mathbf{W})$ is the summation of variance of each block in each diffractive layer. In this case, $\boldsymbol\ell_{r}(\mathbf{W}) = p \times R(\mathbf{W}) + q \times R_{intra}(\mathbf{W})$.

\subsubsection{Smoothing with 2$\pi$}
\label{post_smooth}

\begin{figure*}[!ht]
\centering
  \includegraphics[width=\linewidth]{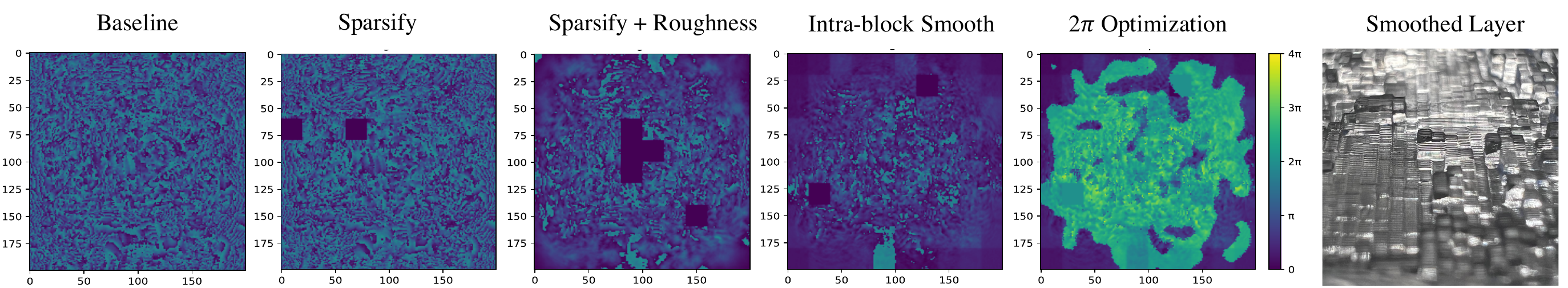}
\caption{Comparison of the phase mask of the second diffractive layer under EMNIST dataset. The black blocks mean weights inside are sparsified and set to zero. The fifth is $2\pi$ optimization of phase mask that trained with sparsification, roughness and intra-block smoothness. The last is 3D printed smoothed diffractive layer. 
}
\label{map}
\vspace{-0.4cm}
\end{figure*}

{
As block sparsified pixels are all zeros while other pixels around the sparsified blocks can have high positive values, which results in larger roughness at the block sparsified pixels and its surrounding pixels as shown in Table \ref{mnist} -- \ref{emnist} and Fig. \ref{map}. Thus, we need post-processing for recovering the roughness.
We observe that the phase modulation of light signal features with 2$\pi$ period, i.e., for a trained phase modulation $c \in [0, 2\pi]$, we have $f(c + 2\pi) = f(c)$, where $f$ is the forward function of the DONN system, which provides us the opportunity to smoothen the phase mask by adjusting the pixels in diffractive layers with $2\pi$ in DONN models. 
Thanks to this characteristic, we can optimize the roughness of the phase mask by selectively adding $2\pi$ to each pixel without retraining as no performance change happens in the model inference. 

Specifically, we formulate the selection of adding $2\pi$
to each pixel in diffractive layers as a combinatorial optimization (CO) problem, and solve it with Gumbel-Softmax (GS)~\cite{jang2016categorical} with gradient descent algorithm. 
In forward, for a phase mask with size of $200\times200$, we have a selection mask of $200\times200\times2$ for selecting the add-on phase of $0$ or $2\pi$ for each pixel in the phase mask, where the selection at each pixel is one-hot represented. Thus, with a matrix of $[[0], [2\pi]]$, by matrix multiplication between $200\times200\times2$ selection mask and the vector $[[0],[2\pi]]$ with size of $2\times1$, the resulting add-on phase mask is $200\times200\times1$ containing either $0$ or $2\pi$ as the add-on phase. The loss function is the roughness computed with the add-on phase and the original phase mask as discussed in Section \ref{roughness}. 

In
backward,
we minimize the roughness loss function. The GS algorithm makes the discrete one-hot represented selection differentiable with continuous approximation of discrete samples~\cite{jang2016categorical}. Specifically, starting from the loss function, it approximates the one-hot selection with continuous probability selection and update probability for the selection options with the backpropagated gradients. Thus, in the next iteration, the loss function is updated with the newly optimized selection probability.
}

\section{Experiment}

\subsection{System Parameters and Training Setup}

{\subsubsection{DONN System Parameters} We design the
system
with three diffractive layers with
the size of layers and the size of total ten detector regions $200 \times 200$. The pixel size is $36~\mu m$ such that the dimension of each fabricated diffractive layer is $720~\mu m \times 720~\mu m$. To fit the optical system, we interpolate the original input images from the MNIST, FMNIST, EMNIST, and KMNIST datasets with size of $28 \times 28$
$200 \times 200$, and encode with the laser source whose wavelength is $532~nm$. The physical distances between layers, first layer to source, and final layer to detector, are set to be $27.94~cm$. 
Separate detector regions for 
different classes are placed evenly on the detector plane with the size of $20 \times 20$, where the sums of the intensity of these regions are equivalent to a
vector in \texttt{float32} type. The final prediction results will be generated using \texttt{argmax}. 
}

\subsubsection{Training Setup} We train the baseline models for  all the four datasets with learning rate being $0.2$ under Adam~\cite{kingma2014adam}, batch-size being $200$. For sparsification, SLR parameters are set to $\rho = 0.1, M = 300,r = 0.1, s_0 = 0.01$. The learning rate is set to $0.001$ with Adam and batch-size $200$. sparsification ratio is set to $0.1$. All implementations are constructed using PyTorch v1.8.1, and results are conducted on Nvidia 2080 Ti GPU.


\subsection{Accuracy and Roughness Evaluation}

We present the results of applying our proposed physics-aware roughness optimization to the four datasets.
In all the experiments, \textbf{Ours-A} refers to the roughness-aware trained model; \textbf{Ours-B} refers to the model trained with sparsity; \textbf{Ours-C} refers to the model trained with sparsity and roughness; \textbf{Ours-D} refers to the model trained with sparsity, plus roughness and intra-block smoothness. The system roughness score is calculated as the average of the roughness of all phase masks: $R_{overall} = \overline{R(\mathbf{W})}$, where $R(\mathbf{W})$ is from Equation~\ref{rough_score}. 
$R_{overall}$ quantifies the overall interpixel interaction within all the phase masks in a DONN system. A lower score means weaker interpixel interaction and less mismatch between numerical modeling and practical deployment.


\begin{table}[t]
\caption{MNIST Result. Baseline is trained under $50$ epochs. The sparsification are trained with block size equal to $25$.}
\label{mnist}
\centering
\begin{adjustbox}{width=\columnwidth, center}
\scriptsize
\begin{tabular}{c|c|cc}
\toprule
\textbf{Model} & 
\multicolumn{1}{c|}{\begin{tabular}[c]{@{}c@{}}{\bf Accuracy} \\ {\bf (\%)}\end{tabular}} & 
\multicolumn{1}{c}{\begin{tabular}[c]{@{}c@{}}{\bf $R_{overall}$ before} \\ {\bf $2\pi$ optimization }\end{tabular}} & 
\multicolumn{1}{c}{\begin{tabular}[c]{@{}c@{}}{\bf $R_{overall}$ after} \\ {\bf $2\pi$ optimization }\end{tabular}} \\
\midrule
\cite{lin2018all, zhou2021large,li2022physics} & 96.67 & 466.39 & 460.85 \\
\midrule
Ours-A & 96.18 & 416.07 & -- \\
Ours-B & 96.38 & 538.78 & 400.38 \\ 
Ours-C & 96.47 & 409.41 & 299.87 \\
Ours-D & 95.90 & 375.35 & \textbf{280.32} \\
\toprule
\end{tabular}
\end{adjustbox}
\end{table}


\begin{table}[t]
\caption{FMNIST Result. Baseline is trained under $150$ epochs. The sparsification are trained with block size equal to $20$.}
\label{fmnist}
\centering
\begin{adjustbox}{width=\columnwidth, center}
\scriptsize
\begin{tabular}{c|c|cc}
\toprule
\textbf{Model} & 
\multicolumn{1}{c|}{\begin{tabular}[c]{@{}c@{}}{\bf Accuracy} \\ {\bf (\%)}\end{tabular}} & 
\multicolumn{1}{c}{\begin{tabular}[c]{@{}c@{}}{\bf $R_{overall}$ before} \\ {\bf $2\pi$ optimization }\end{tabular}} & 
\multicolumn{1}{c}{\begin{tabular}[c]{@{}c@{}}{\bf $R_{overall}$ after} \\ {\bf $2\pi$ optimization }\end{tabular}} \\
\midrule
\cite{lin2018all,zhou2021large,li2022physics} & 87.98 & 464.78 & 461.98 \\
\midrule
Ours-A & 86.99 & 421.49 & -- \\
Ours-B & 87.88 & 488.11 & 438.53 \\ 
Ours-C & 86.79 & 350.67 & 305.86 \\
Ours-D & 85.76 & 450.73 & \textbf{229.70} \\
\toprule
\end{tabular}
\end{adjustbox}
\end{table}


\begin{table}[t]
\caption{KMNIST Result. Baseline is trained under $100$ epochs. The sparsification are trained with block size equal to $20$.}
\label{kmnist}
\centering
\begin{adjustbox}{width=\columnwidth, center}
\scriptsize
\begin{tabular}{c|c|cc}
\toprule
\textbf{Model} & 
\multicolumn{1}{c|}{\begin{tabular}[c]{@{}c@{}}{\bf Accuracy} \\ {\bf (\%)}\end{tabular}} & 
\multicolumn{1}{c}{\begin{tabular}[c]{@{}c@{}}{\bf $R_{overall}$ before} \\ {\bf $2\pi$ optimization }\end{tabular}} & 
\multicolumn{1}{c}{\begin{tabular}[c]{@{}c@{}}{\bf $R_{overall}$ after} \\ {\bf $2\pi$ optimization }\end{tabular}} \\
\midrule
\cite{lin2018all,zhou2021large,li2022physics} & 86.92 & 460.61 & 445.57 \\
\midrule
Ours-A & 85.26 & 462.7 & -- \\
Ours-B & 86.83 & 473.08 & 432.26 \\ 
Ours-C & 85.01 & 396.84 & 331.22 \\
Ours-D & 83.19 & 327.48 & \textbf{288.42} \\
\toprule
\end{tabular}
\end{adjustbox}
\end{table}


\begin{table}[t]
\caption{EMNIST Result. Baseline is trained under $100$ epochs. The sparsification are trained with block size equal to $20$.}
\label{emnist}
\centering
\begin{adjustbox}{width=\columnwidth, center}
\scriptsize
\begin{tabular}{c|c|cc}
\toprule
\textbf{Model} & 
\multicolumn{1}{c|}{\begin{tabular}[c]{@{}c@{}}{\bf Accuracy} \\ {\bf (\%)}\end{tabular}} & 
\multicolumn{1}{c}{\begin{tabular}[c]{@{}c@{}}{\bf $R_{overall}$ before} \\ {\bf $2\pi$ optimization }\end{tabular}} & 
\multicolumn{1}{c}{\begin{tabular}[c]{@{}c@{}}{\bf $R_{overall}$ after} \\ {\bf $2\pi$ optimization }\end{tabular}} \\
\midrule
\cite{lin2018all,zhou2021large,li2022physics} & 92.30 & 463.42 & 458.48 \\
\midrule
Ours-A & 91.61 & 435.58 & -- \\
Ours-B & 92.36 & 465.85 & 443.91 \\ 
Ours-C & 91.16 & 349.61 & 336.75 \\
Ours-D & 90.74 & 312.17 & \textbf{298.09} \\
\toprule
\end{tabular}
\end{adjustbox}
\end{table}


Table~\ref{mnist}, \ref{fmnist}, \ref{kmnist} and \ref{emnist} compare the accuracy and roughness score of each component of our proposed optimization method on the four datasets.
For all the experiments,  \cite{lin2018all} has the highest roughness score.
With the $2\pi$ optimization, the roughness score dropped less than $2\%$, which means the phase change between pixels in the phase masks is very drastic within $2\pi$ range. 
The roughness-aware training (Ours-A) results to around $10\%$ roughness drop but still with relatively high roughness score. The sparsification (Ours-B) results to higher accuracy compared with roughness-aware training but also higher roughness score because of the sharp phase changes between sparsified pixels and their surroundings as explained in Section~\ref{post_smooth}. With the help of $2\pi$ optimization, the phase gap is smoothed with the $2\pi$ periodic feature and the roughness score dropped to the same level as of roughness-aware training. 
Hence, the combination of the sparsification and the roughness modeling (Ours-C) shows great improvement. It results to the drop of the roughness score in $12.2\%$, $24.6\%$, $13.8\%$ and $24.6\%$ respectively for the four datasets before the $2\pi$ optimization, $35.7\%$, $34.2\%$, $28.1\%$ and $27.3\%$ after the $2\pi$ optimization, with little impact on accuracy.
The introduction of intra-block smoothness (Ours-D) further reduces the roughness score. Given a bit more accuracy flexibility (in average 2\% acorss all datasets), we reduce the roughness after $2\pi$ optimization by $40\%$, $50.6\%$, $37.4\%$ and $35.7\%$ for the four datasets, respectively.


Fig.~\ref{map} compares the second phase mask (diffractive layer) of DONN in different models under the EMNIST dataset. 
The first four are phase masks without the $2\pi$ optimization. The black blocks are the sparsified areas,
which form sharp contrast with the surrounding colors. The last is the phase mask from the model trained with sparsity, roughness and intra-block smoothness and with $2\pi$ post-smooth optimization applied.
With the selectively addition of $2\pi$ to each pixel in the entire phase mask, the previous black areas blend in with the surrounding color and the entire masks become smoother.
Moreover, with the awareness of the mask roughness during training, the valid phase modulations in diffractive layers are more clustered at the center part of the mask while the edge part is more likely to provide marginal modulations.
This correlates with the optical characteristics where the light signal strength decays from the center. Thus, the most valid information and phase modulation are expected to happen at the center of the mask. The smoothed phase mask will result in easy fabrication of the mask and better correlation for DONN numerical modelling and its physical deployment.


\subsection{Design Exploration}

\begin{figure}[t]
\centering
  \includegraphics[width=\linewidth]{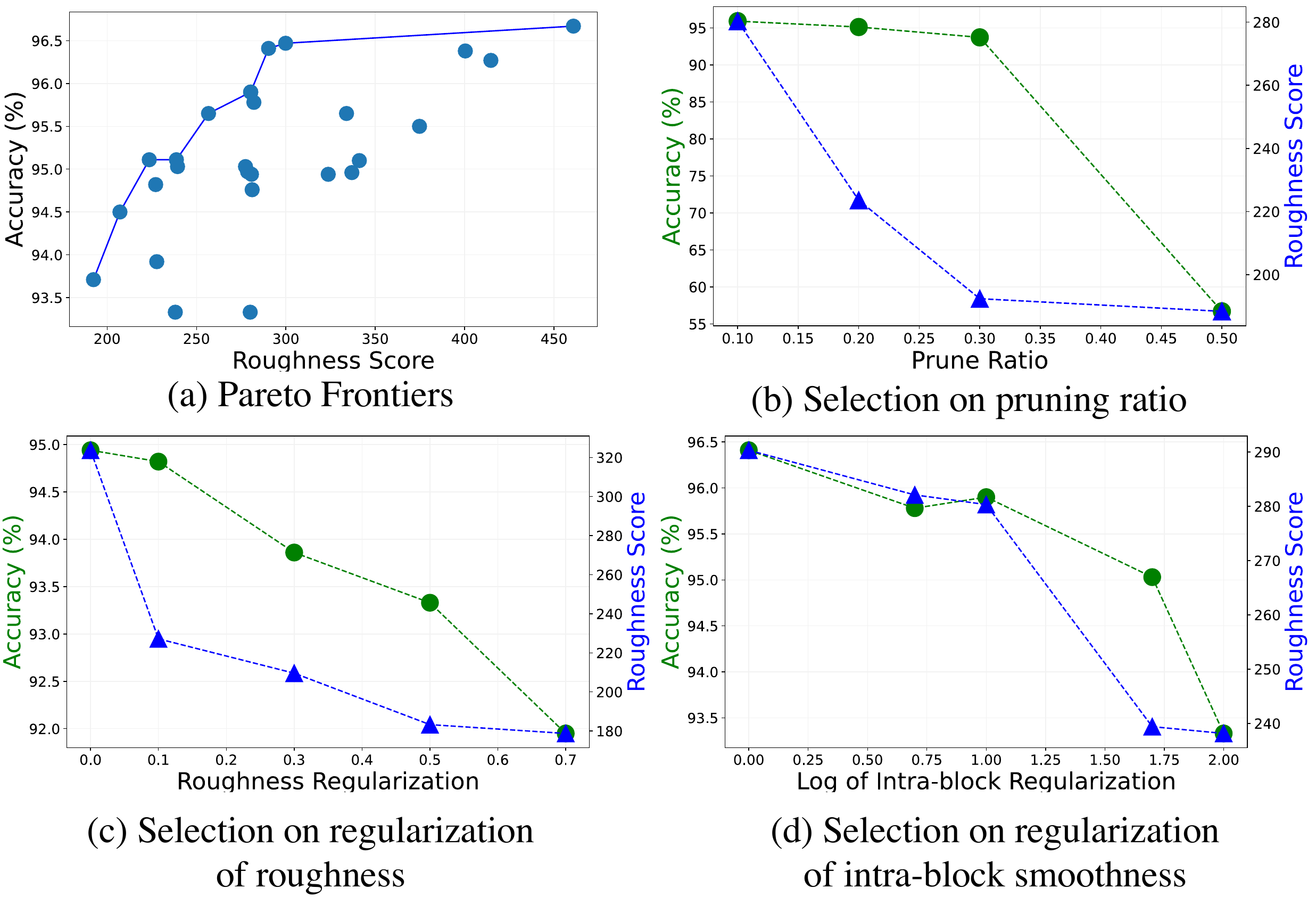}
\vspace{-0.7cm}
\caption{Hyperparameter Exploration. (a) Pareto frontiers of accuracy vs. roughness score of explorations on MNIST. (b) - (d) Exploration of sparsification ratio, regularization of roughness and intra-block smoothness vs. accuracy and roughness score. 
}
\label{DE_plot}
\end{figure}

We 
further perform
hyperparameter exploration in Fig.~\ref{DE_plot}. Fig.~\ref{DE_plot} (a) shows the Pareto frontier for roughness score vs. accuracy. It is clear that as the accuracy increases, the roughness score increases accordingly, which means the performance mismatch between digital emulation and hardware deployment of DONN system would be larger. In this case, hyperparameters need to be adjusted to address this trade-off. For Fig.~\ref{DE_plot}(b)-(d), we further explore the relation of accuracy and roughness score vs. sparsification ratio and regularizations of roughness and intra-block smoothness. We observe that accuracy and roughness score both decrease but under different magnitudes as the sparsification ratio and regularization increase. For roughness regularization in Fig.~\ref{DE_plot}(c), both accuracy and roughness score show an inflection point at $0.1$, with accuracy subsequently showing a rapid decrease, and roughness becomes smoother. Similarly for intra-block regularization in Fig.~\ref{DE_plot}(d), the inflection point is shown when the $\log$ of regularization is $1$. We also observe that these trends hold for the other three datasets.

\vspace{-0.1cm}
\section{Conclusion}

In this paper, we propose a physics-aware roughness optimization framework for diffractive optical neural networks, aiming to narrow the performance mismatch when deploying the digitally emulated DONN to practical optical devices. We introduce sparsity into phase masks through block sparsification and integrates roughness regularization into DONN loss function to reduce the interpixel interaction within diffractive layers. $2\pi$ periodic phase modulation and intra-block smoothness are applied for further smoothness. Results show that our physics-aware roughness optimization can provide $35.7\%$, $34.2\%$, $28.1\%$, and $27.3\%$ reduction in roughness with only minor accuracy loss on MNIST, FMNIST, KMNIST, and EMNIST, respectively. Given a bit more accuracy flexibility using the proposed intra-block smoothness, we could further reduce the roughness by $40\%$, $50.6\%$, $37.4\%$, and $35.7\%$.


\section*{Acknowledgment}

This work is in part supported by National Science Foundation (NSF), under NSF-2047176, NSF-2008144, and NSF-2019336 awards.

\tiny
\bibliographystyle{unsrt}
\bibliography{ref}

\end{document}